# Assisted Probe Positioning for Ultrasound Guided Radiotherapy Using Image Sequence Classification

Alex Grimwood[1,2*[0000-0002-2608-2580]], Helen McNair[1[0000-0001-7389-5587]], Yipeng Hu[2,3[0000-0003-4902-0486]], Ester Bonmati[2[0000-0001-9217-5438]], Dean Barratt[2,3[0000-0003-2916-655X]], Emma J. Harris[1[0000-0001-8297-0382]]

[1] Division of Radiotherapy and Imaging, Institute of Cancer Research, Sutton, UK
[2] Wellcome/EPSRC Centre for Interventional and Surgical Sciences, University College London, London, UK
[3] Centre for Medical Image Computing, University College London, London, UK
alex.grimwood@ucl.ac.uk

**Abstract.** Effective transperineal ultrasound image guidance in prostate external beam radiotherapy requires consistent alignment between probe and prostate at each session during patient set-up. Probe placement and ultrasound image interpretation are manual tasks contingent upon operator skill, leading to interoperator uncertainties that degrade radiotherapy precision. We demonstrate a method for ensuring accurate probe placement through joint classification of images and probe position data. Using a multi-input multi-task algorithm, spatial coordinate data from an optically tracked ultrasound probe is combined with an image classifier using a recurrent neural network to generate two sets of predictions in real-time. The first set identifies relevant prostate anatomy visible in the field of view using the classes: outside prostate, prostate periphery, prostate centre. The second set recommends a probe angular adjustment to achieve alignment between the probe and prostate centre with the classes: move left, move right, stop. The algorithm was trained and tested on 9,743 clinical images from 61 treatment sessions across 32 patients. We evaluated classification accuracy against class labels derived from three experienced observers at 2/3 and 3/3 agreement thresholds. For images with unanimous consensus between observers, anatomical classification accuracy was 97.2% and probe adjustment accuracy was 94.9%. The algorithm identified optimal probe alignment within a mean (standard deviation) range of 3.7° (1.2°) from angle labels with full observer consensus, comparable to the 2.8° (2.6°) mean interobserver range. We propose such an algorithm could assist radiotherapy practitioners with limited experience of ultrasound image interpretation by providing effective real-time feedback during patient set-up.

## 1 Introduction

Advanced external beam radiotherapy techniques, such as stereotactic body radiotherapy, adaptive radiotherapy and ultrahypofractionation require high precision spatiotemporal image guidance [1-3]. Ultrasound imaging is a desirable guidance technology due to the modality's excellent soft tissue contrast, high spatial resolution and ability to scan in both real-time and three dimensions. Transperineal ultrasound (TPUS) is used clinically to assist with the localization and delineation of anatomical structures for prostate radiotherapy, showing comparable precision to standard cone beam computed tomography scans (CBCT) [4]. The efficacy of ultrasound guided radiotherapy is heavily reliant upon experienced staff with a level of sonography training not typically acquired in the radiotherapy clinic [5]. With this in mind, computational technologies are being developed to assist TPUS operators in radiotherapy so that they can achieve the levels of precision demanded by state-of-the-art treatment delivery methods.

Radiotherapy patients require multiple treatment sessions. To achieve effective image guidance, anatomical structures such as the prostate must be precisely localized and registered to a planning scan at each session. Automated registration methods have been investigated as a way to localize the prostate prior to treatment [6]. Such methods neglect the necessary step of ensuring probe placement is consistent with the planning scan, which can affect radiotherapy accuracy [7]. Previous studies have reported automatic identification of the optimal probe position on CT scans, however the method is computationally intensive and requires good spatial agreement between the planning CT scan and the patient many weeks later during set-up [8].

In this study, we demonstrate a method for identifying the optimal probe position by presenting the operator with actionable information in near real-time using only ultrasound images and probe tracking data obtained at patient set-up. Our method uses a supervised deep learning approach to recommend probe adjustments in response to image content and probe position.

Deep learning methods have been used extensively in other ultrasound applications [9]. Automatic segmentation of pelvic structures, such as the prostate and levator hiatus, has demonstrated performance levels exceeding manual contouring [10, 11]. Real-time classification and localization of standard scan planes in obstetric ultrasound has been demonstrated [12]. Image guidance in 3D Ultrasound target localization has also been reported for epidural needle guidance [13].

For this study, a multi-input multi-task classification approach was adopted to utilize data acquired routinely during probe set-up. The input data comprised 2D (B-scan) TPUS images and spatial coordinate information for each scan plane obtained from optical tracking. The classification labels were designed to be easily interpreted, presenting the user with simple information regarding image content and recommended probe adjustments.

## 2 Methods

In this section, the collection and labelling of image data is described along with the design and implementation of the classification models used in the study. The training and evaluation methods against three expert observers is also described.

### 2.1 Data Collection and Labelling

Clinical data was acquired using the Clarity Autoscan™ system (Elekta AB, Stockholm.) as part of the Clarity Pro trial (NCT02388308) approved by the Surrey and SE Coast Region Ethics Committee and described elsewhere [14]. The system incorporated an optically tracked TPUS probe with a mechanically swept transducer array, which was used to collect 3D prostate ultrasound scans during patient set-up. According to the trial protocol, a trained radiotherapy practitioner used the Clarity probe to localize the central sagittal plane through the prostate with real-time freehand 2D imaging. Once identified, the probe was clamped in place and a static volumetric scan was acquired. Volumetric acquisition comprised recording a sequence of 2D images (B-scans) while mechanically sweeping the transducer through a 60° arc from right to left in the patient frame of reference. Each B-scan from the volume was automatically labelled with the imaging plane location and orientation in room coordinates. A total of 9,743 B-scans were acquired comprising 61 volumes across 32 patients. Images were 640×480 pixels, with pixel sizes between 0.3 mm to 0.5 mm and volumes comprising 134 to 164 B-scans depending on scan parameters.

All scan volumes were reviewed and the constituent B-scans assigned position labels by three experienced observers (2 physicists and 1 radiotherapy practitioner) as either: Centre (C) for image planes through the prostate centre; Periphery (P) for image planes through the outer prostate; Outside (O) for image planes not intersecting the prostate. The prostate centre was distinguishable from the periphery by prostate shape and anatomical features such as the urethra branching from the penile bulb. Direction labels were automatically assigned depending on the transducer motion required to position the B-scan plane at the prostate centre, as ensured manually during the acquisition. Centre images were labelled Stop (S), while Periphery and Outside images were labelled as either Left (L) or Right (R) depending on which side of the prostate centre they were located to indicate the direction of movement needed to align the transducer with the prostate. Position and direction input labels were treated as two separate class sets. Each set was encoded as an ordinal 3-vector of probabilities derived from observer consensus in each class, where: 100% consensus = 1, 66.7% (2/3) consensus = 0.667, 33.3% (1/3) consensus = 0.333.

### 2.2 Classifier Models

Two classifier models were developed, referred to as: MobNet and RNN (Figure 1). MobNet, was an image classifier based upon the Keras implementation of MobileNetV2 pre-trained on the ImageNet dataset [15]. Single B-scans images were passed to the convolutional network and a custom fully connected layer was added to concatenate the last convolutional pooling layer with a 6-tuple input representing the image plane coordinates. Two dense branches each produced three class outputs. The position branch classified images according to visible prostate anatomy: Outside (O), Periphery (P) and Centre (C). The direction branch classified images according to the transducer rotation direction required to align the image plane with the prostate centre: Right (R), Stop (S) and Left (L).

The second model, RNN, extended the MobNet. To improve classification of a given B-scan, $P_0$, the image was grouped with 9 immediately preceding B-scans from the volumetric sweep to form a 10-image sequence $P_{n-10}$ ($n$ = 1, ..., 10). Each image was processed individually by the same MobileNetV2 backbone used for MobNet. The output from these convolutional layers was concatenated with probe orientation data and fed into the recurrent layers in sequence. The recurrent position and direction branches comprised long-short-term memory (LSTM) layers, sharing between the images to predict a single output for each 10-image sequence.

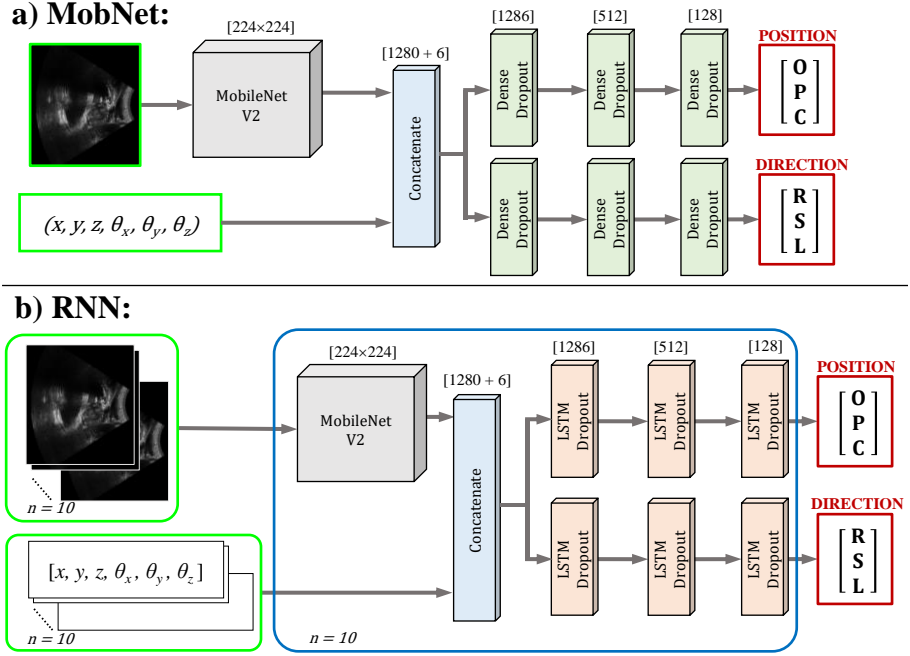

**Fig. 1.** MobNet (a): inputs in green are a 2D B-scan and transducer position (x, y, z) with orientation ($\theta_x$, $\theta_y$, $\theta_z$). Outputs are two class sets in red: Outside, Periphery, Centre (O, P, C) for position and Right, Stop, Left (R, S, L) for direction. Input dimensions are shown in square brackets for MobileNetV2, concatenation, dense and LSTM layers. RNN (b): input sequences are generated from 10 consecutive images and transducer positions. Each is concatenated (blue box) and passed onto the LSTM layers.

### 2.3 Experiments

A training set comprising 8,594 images (54 volumes, 27 patients) was used for MobNet. 8,108 image sequences were produced from the same training set for RNN. A 4-fold cross validation was used for both models over 100 epochs with an Adam optimizer (learning rate = $1\times10^{-4}$). Class imbalances were compensated for using sample weighting where label proportions were: [O 62%, P 23%, C 15%] and [R 43%, S 15%, L 42%]. Images were downsampled by a factor of 1.5, normalised and cropped to a fixed 224×224 pixel region of interest large enough to encompass the prostate where present.

Image augmentation was applied. Augmentation steps included random rotations up to ±45° and random translations up to ±20 pixels, approximating the range of clinically observed variation between individual sessions and patients. Random flips along both vertical and horizontal axes were also applied.

The trained models were assessed on a separate test set of 1,149 images (7 volumes, 5 patients) for MobNet, equating to 1,086 sequences for RNN. Image quality was assessed for every volume by a single observer. A subjective score (0 to 3) was assigned depending on visible anatomical features: prostate boundaries, seminal vesicles, penile bulb, urethra and contrast (Figure 2). The proportion of scores was maintained between training and test datasets.

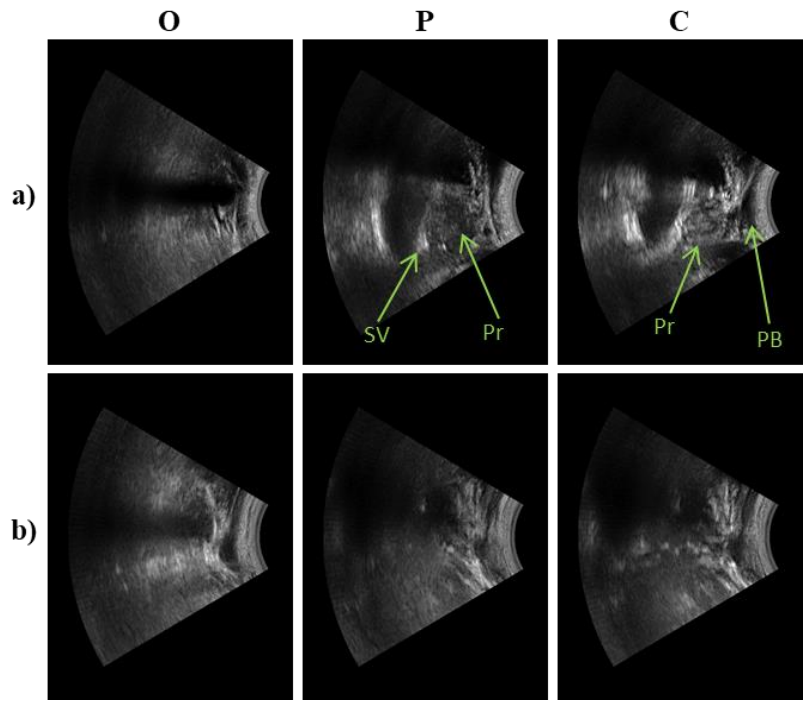

**Fig. 2.** Sagittal B-mode ultrasound images outside of the prostate (O), within the lateral prostate periphery (P) and through the prostate centre (C). B-scans are shown from a volume with an image quality score of 3 (a) and score of 0 (b), determined by visible anatomical features, such as: seminal vesicles (SV), penile bulb (PB) and prostate with well-defined boundary (Pr).

### 2.4 Analysis

MobNet and RNN model output was assessed against the observer labels for: 1) a threshold of 66.7%, where a positive label was defined as ≥2/3 consensus; and 2) a 100% consensus threshold, where only samples with full consensus were assessed.

Precision, recall, accuracy and F1-scores were calculated and confusion matrices plotted. RNN and MobNet accuracies were compared using McNemar tests, which are analogous to paired t-tests for nominal data [16]. Agreement between RNN and the observer cohort was quantified by calculating the Williams Index 95% confidence intervals [17]. Interobserver agreement was also quantified by calculating Fleiss' Kappa and Specific Agreement coefficients for each class [18, 19]. Interobserver uncertainty of the transducer angles associated with the prostate centre was quantified by calculating the mean difference between each observer and the 100% consensus labels for the 7 test volumes. RNN angular uncertainty was also calculated and compared.

## 3  Results and Discussion

RNN classification performance is shown in Table 1 along with the associated confusion matrices in Figure 3. Accuracy was highest among images with 100% consensus labels, as was specificity, recall and F1-score. Maximum position and direction accuracies were 97.2% and 94.9% respectively.

Accuracy, F1-score, precision and recall were consistently higher for RNN than MobNet (Supplementary Materials). McNemar comparison tests indicated the RNN accuracy improvement over MobNet was significant for position classification (P < 0.001), but only marginal for direction classification (P = 0.0049 at 66.7%, P = 0.522 at 100%).

Agreement between model output and the observer cohort was quantified using Williams indices at 95% confidence intervals (CI). RNN position output significantly outperformed the observer cohort, having CI bounds between 1.03 and 1.05. RNN direction output exhibited poorer agreement, with a CI between 0.95 and 0.98; however, this is unsurprising because direction labels were assigned using information from the entire volumetric scan, rather than a 10-image sub-volume.

Fleiss' Kappa scores indicated excellent interobserver agreement (>0.8) among image labels for the position class set, with good agreement (>0.7) for the direction class set. Specific agreement coefficients indicated consistently high interobserver agreement (>0.8) within classes (full figures provided in Supplementary Materials). For the test dataset 100% position consensus was achieved for 884 images (821 sequences) and 100% direction consensus was achieved for 1,036 images (973 sequences). All images and sequences achieved at least 66.7% consensus in both class sets.

Table 1. RNN precision (prec), recall (rec), F1-score (F1) and accuracy for both 100% (full) and 66.7% (2/3) observer consensus label thresholds.

| Class Set | Class Label | Full Consensus Threshold | | | 2/3 Consensus Threshold | | |
|---|---|---|---|---|---|---|---|
| | | Prec % | Rec % | F1 % | Prec % | Rec % | F1 % |
| Position | Outside | 99.1 | 98.7 | 98.9 | 95.6 | 93.1 | 94.3 |
| | Periphery | 91.6 | 97.5 | 85.7 | 81.9 | 85.5 | 83.6 |
| | Centre | 100.0 | 85.7 | 92.3 | 79.5 | 83.6 | 79.8 |
| Direction | Right | 91.1 | 100.0 | 95.3 | 89.6 | 98.8 | 94.0 |

| | | | | | | |
|---|---|---|---|---|---|---|
| Stop | 93.9 | 74.7 | 83.2 | 77.0 | 81.7 | 79.3 |
| Left | 98.9 | 93.9 | 96.3 | 98.8 | 89.5 | 93.2 |
| Position Accuracy % | | 97.2 | | | 89.3 | |
| Direction Accuracy % | | 94.9 | | | 92.4 | |

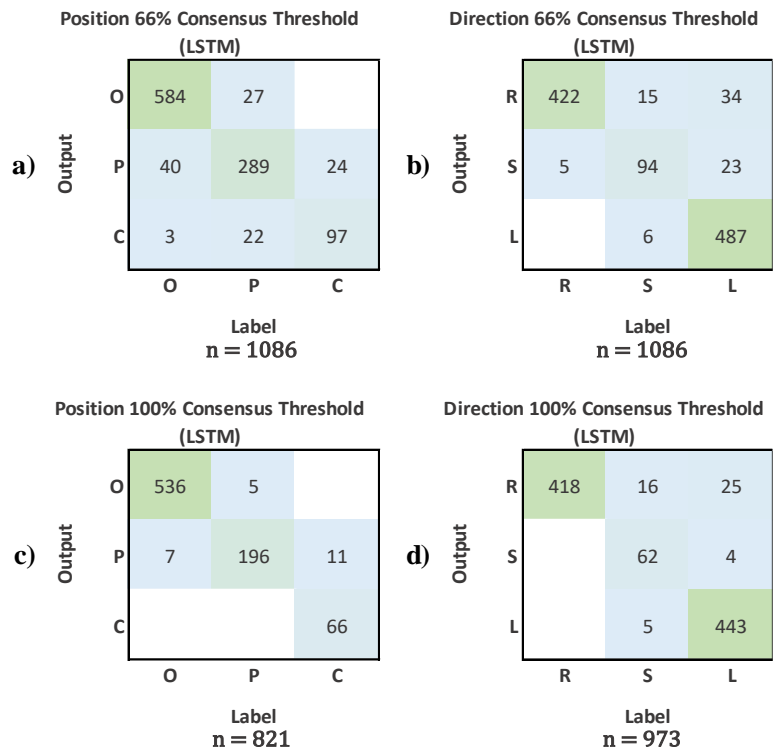

**Fig. 3.** RNN confusion matrices for 66.7% observer consensus thresholds (a, b) and 100% observer consensus thresholds (c, d). Total number of samples in each matrix is given by *n*.

Mean interobserver angular uncertainty (standard deviation) was calculated to be 2.8° (2.6°). A comparable mean uncertainty of 3.7° (1.2°) was calculated for the RNN model. Finally, the RNN mean classification time was 0.24 ms per sequence on a 2.8 GHz Intel Xeon E3 CPU, establishing the possibility of incorporating a practical real-time solution on clinical systems.

## 4      Conclusion

A deep learning classifier incorporating recurrent layers has been shown to predict prostate anatomy and recommend probe position adjustments with high precision and accuracies comparable to expert observers. The recurrent network was significantly more accurate than a non-recurrent equivalent. This study demonstrates the possibility of enhancing ultrasound guidance precision by reducing interobserver variation and assisting ultrasound operators with finding the optimal probe position during patient set-up for ultrasound guided radiotherapy.

**Acknowledgements.** This work was supported by NHS funding to the NIHR Biomedical Research Centre at The Royal Marsden and The Institute of Cancer Research. The study was also supported by Cancer Research UK under Programmes C33589/A19727 and C20892/A23557, and by the Wellcome/EPSRC Centre for Interventional and Surgical Sciences (203145Z/16/Z). The study was jointly supervised by Dr Emma J. Harris, Prof. Dean Barratt and Dr. Ester Bonmati. We thank the radiographers of the Royal Marsden Hospital for their clinical support, as well as David Cooper, Martin Lachaine and David Ash at Elekta for their technical support.